\documentclass{article}
\usepackage{bbm}
\usepackage{spconf,amsmath,graphicx}
\usepackage[abs]{overpic}
\usepackage{cite}
\usepackage{amsmath,amssymb,amsfonts}
\usepackage{amsthm}
\usepackage{dsfont}
\usepackage{algorithmic}
\usepackage{hyperref}
\usepackage{textcomp}
\usepackage{xcolor}
\usepackage{fix-cm}

\usepackage{wrapfig, blindtext}

\usepackage{algorithm}
\usepackage{algorithmic}

\def\BibTeX{{\rm B\kern-.05em{\sc i\kern-.025em b}\kern-.08em
    T\kern-.1667em\lower.7ex\hbox{E}\kern-.125emX}}

\DeclareMathSizes{10}{9}{7}{5}

\begin{document}
\raggedbottom

\title{Semantic-Aware UAV Command and Control for Efficient IoT Data Collection}
 

\name{%
  Assane Sankara \qquad Daniel Bonilla Licea \qquad Hajar El Hammouti \thanks{This document has been produced with the financial assistance of the European Union (Grant no. DCI-PANAF/2020/420-028), through the African Research Initiative for Scientific Excellence (ARISE), pilot programme. ARISE is implemented by the African Academy of Sciences with support from the European Commission and the African Union Commission. The contents of this document are the sole responsibility of the author(s) and can under no circumstances be regarded as reflecting the position of the European Union, the African Academy of Sciences, and the African Union Commission.}
}
\address{ College of Computing, Mohammed VI Polytechnic University (UM6P), Ben Guerir, Morocco \\emails: \{assane.sankara, daniel.bonilla,hajar.elhammouti\}@um6p.ma }

\maketitle

\begin{abstract}
    Unmanned Aerial Vehicles (UAVs) have emerged as a key enabler technology for data collection from Internet of Things (IoT) devices. However, effective data collection is challenged by resource constraints and the need for real-time decision-making. In this work, we propose a novel framework that integrates semantic communication with UAV command-and-control (C\&C) to enable efficient image data collection from IoT devices. Each device uses Deep Joint Source-Channel Coding (DeepJSCC) to generate a compact semantic latent representation of its image to enable image reconstruction even under partial transmission. A base station (BS) controls the UAV's trajectory by transmitting acceleration commands. The objective is to maximize the average quality of reconstructed images by maintaining proximity to each device for a sufficient duration within a fixed time horizon. To address the challenging trade-off and account for delayed C\&C signals, we model the problem as a Markov Decision Process and propose a Double Deep Q-Learning (DDQN)-based adaptive flight policy. Simulation results show that our approach outperforms baseline methods such as greedy and traveling salesman algorithms, in both device coverage and semantic reconstruction quality.
\end{abstract}


\section{Introduction}

Unmanned Aerial Vehicles (UAVs) have emerged as a key technology for extending coverage and collecting data in Internet of Things (IoT) networks, particularly in scenarios where terrestrial communication infrastructure is unavailable or limited \cite{ndiaye2022age,ndiaye2023muti, licea2025reshaping}. Their rapid deployability makes them highly valuable in applications such as precision agriculture and disaster management~\cite{survey_uav_data_collection}. However, IoT devices often generate massive volumes of raw data, and transmitting this data over wireless channels is constrained by the limited communication, energy, and bandwidth resources available to UAVs.

In this context, semantic communication is gaining attention as a promising paradigm to enhance UAV-enabled communications. Unlike traditional communication approaches that focus on transmitting bits accurately, semantic communication aims to convey only the most meaningful and task-relevant information, reducing resource consumption \cite{marnissi2024semantic,chaccour2022dataknowledgebuildinggeneration, frombitstosemantics}.
In the area of image transmission, Deep Joint Source-Channel Coding (DeepJSCC) has been proposed to encode images directly into channel symbols using deep neural networks \cite{Bourtsoulatze_2019,xu2023deepjointsourcechannelcoding}. Furthermore, a few works have explored the integration of generative models into semantic communication frameworks \cite{liu2024diffusionbasedgenerativemulticastingintentaware, qiao2024latencyawaregenerativesemanticcommunications, xu2024semanticawarepowerallocationgenerative}. In these works, textual prompts are first extracted from images, transmitted using resource-efficient approaches, and then used to regenerate high-quality images at the receiver through diffusion models. 
Beyond image transmission, semantic communication has shown great potential in enhancing UAV command and control (C\&C) systems. In \cite{goal_orient_sem_com_uav}, the UAV leverages metrics such as the age of information (AoI) and the value of information (VoI) to assess the semantic importance of the command and control data. In \cite{xu2023taskorientedsemanticsawarecommunicationwireless}, the framework considers both the similarity and the freshness of successive command and control data to avoid transmitting redundant information.

The aforementioned works deal with the optimization of UAV C\&C messages separately from the semantic transmission of IoT data. Moreover, most existing approaches assume ideal, delay-free command transmission and overlook the trade-off between the optimization of the UAV's trajectory 
and the need to ensure high-quality semantic data recovery. 
In this paper, we address this gap by making three key contributions. First, we design a semantic-aware data collection framework in which a UAV travels from an initial location to a destination while dynamically adapting its trajectory to maximize the average quality of the reconstructed images. Second, to address the studied problem, we formulate it as a Markov Decision Process (MDP) and propose a Double Deep Q-Learning (DDQN)-based control algorithm to control the UAV trajectory and successfully deliver the semantic information. Third,  through simulation experiments, we show that our approach learns an adaptive flight policy and outperforms baseline methods in terms of both number of visited devices and semantic reconstruction quality.


\section{System Model}


We consider a set of $N$ IoT devices, denoted by $\mathcal{N} = \{1, 2, \dots, N\}$, deployed within a given area. Each IoT device $n \in \mathcal{N}$ is located at a fixed position $\mathbf{q}_n^{\text{IoT}} = (x_n^{\text{IoT}}, y_n^{\text{IoT}}, 0)$. 
A UAV departs from a starting point, collects data from these devices during its flight, and reaches a destination within a time horizon $T$. Data collection occurs while the UAV is in motion: transmission starts when a device enters the UAV's communication range $D^\text{com}$, and it continues until the device falls out of range \footnote{This setup reflects practical scenarios in which a UAV is required to travel from a source to a destination and has to adapt its trajectory to opportunistically collect data from ground devices during its flight.}. 

\begin{figure}[t]
    \centering
    \includegraphics[width=\linewidth]{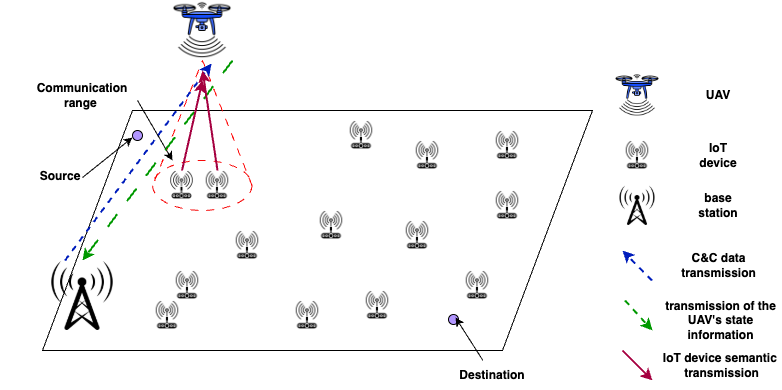}
    \vspace{-0.5cm}
  \caption{System model.}
    \label{fig:system_model}
\end{figure}

The UAV's trajectory is controlled through C\&C messages sent by a BS located at coordinates $\mathbf{q}^{\text{BS}}=(x^{\text{BS}}, y^{\text{BS}}, h^{\text{BS}})$, where $h^{\text{BS}}$ denotes the BS's height, and $(x^{\text{BS}}, y^{\text{BS}}) \in \mathbb{R}^2$ is its 2-dimensional position.
Time is discretized into $K$ slots, each of duration $\tau$, such that the mission duration is $T=K\tau, K \in \mathbb{N}$. At the end of each time slot $t \in \{1,..,K\}$, the UAV's location is denoted by $\mathbf{q}^{\text{UAV}}[t] = (x^{\text{UAV}}[t], y^{\text{UAV}}[t], h^{\text{UAV}})$, where $h^{\text{UAV}}$ is the UAV's fixed flying altitude. Figure \ref{fig:system_model}
illustrates the system model.

Based on the UAV's position and velocity, $\mathbf{q}^{\text{UAV}}[t]$ and $\mathbf{v}[t]$ at the end of time interval $t$, the BS computes a new 2D acceleration vector $\mathbf{a}[t+1] = (a^x[t+1], a^y[t+1])$, where $a^x[t+1]$ and $a^y[t+1]$ represent the acceleration components along the 
$x$- and $y$-axes for the upcoming interval $t+1$.

During the horizon $T$, we assume that each IoT device $n$ has an image $\mathbf{I}_n$ to transmit \footnote{Note that data generation and queuing are beyond the scope of this paper.}. To enable efficient communication, each device processes its image using a DeepJSCC encoder \cite{xu2023deepjointsourcechannelcoding}, producing a compressed latent representation composed of complex-valued symbols. Specifically, the encoder produces a vector $\mathbf{z}_n = \{z^1_n, z^2_n, ..., z^M_n\}$, where $z^m_n$ (with $m \in \{1, 2, ..., M\}$) is the $m$-th complex symbol of the latent vector and $M$ is the number of complex symbols in the latent vector. On the UAV side, the DeepJSCC decoder reconstructs the image from the received symbol vector $\mathbf{\hat{z}}_n$ to obtain a decoded image $\hat{\mathbf{I}}_n$. 
Following the training approach described in \cite{kurka2021bandwidthagileimagetransmissiondeep}, the latent vector $\mathbf{z}_n$ is structured such that the information is ordered in descending importance (i.e., from the symbol with the highest magnitude to the lowest). As a result, even if only a partial set of symbols is received, the decoder can still attempt reconstruction by padding the missing symbols with zeros. The more symbols received, the better the quality of the reconstructed image \cite{kurka2021bandwidthagileimagetransmissiondeep}.

During data collection, when an IoT device is within the range of the UAV, it starts transmitting its symbols to the UAV. Transmission continues as long as the UAV remains within the communication range. Once the UAV exits this range, the IoT device stops transmitting (even if it has not sent all its symbols). Consequently, some IoT devices may be unable to transmit their full latent vector. However, image reconstruction is still possible using the subset of symbols that were successfully received. When multiple IoT devices simultaneously fall within the UAV's communication range, they share the channel using Orthogonal Frequency-Division Multiple Access (OFDMA).
Given this scenario, the BS's objective is to optimize the UAV's acceleration commands so that the UAV can visit as many IoT devices as possible while staying within range long enough to collect the maximum number of symbols from each. Maximizing the number of received symbols directly improves the quality of the reconstructed images.
The reconstruction quality is quantified using the Peak Signal-to-Noise Ratio (PSNR) \cite{xu2023deepjointsourcechannelcoding}, computed between the original image $\mathbf{I}_n$ and the reconstructed image $\hat{\mathbf{I}}_n$ as follows:
\vspace{-0.09cm}
\begin{equation}
    \phi_n(\mathbf{I}_n, \hat{\mathbf{I}}_n) = 10 \log_{10}\Bigl( \zeta_{\text{max}}^2 / \text{MSE}(\mathbf{I}_n, \hat{\mathbf{I}}_n)\Bigr),
\end{equation} where $\zeta_{\text{max}}$ is the maximum possible value of the image pixels\footnote{For example, for images with pixel intensities in the range [0, 255], $\zeta_{\text{max}}$ is defined as 255, while for images normalized to the range  [0, 1], $\zeta_{\text{max}}$ is 1.}. MSE is the mean squared error between the two images.


We adopt an air-to-ground channel model to characterize the communication links between the UAV and both the BS and the IoT devices. We are interested in the transmissions from the BS to the UAV, and from IoT devices to the UAV. Let $n \in \mathcal{N} \cup \{b\}$ denote a generic node, where $b$ is the index corresponding to the BS. The achievable transmission rate from node $n$ to the UAV at time $t$ is given by:
\vspace{-0.1cm}
\begin{equation}
    R_{n}[t] = B_n\log_2\left(1 + \frac{P_n}{\sigma^2 \, 10^{\text{PL}_n[t]/10}}\right),
\end{equation} where $B_n$ is the bandwidth allocated to node $n$. $P_n$ is the transmit power of node $n$ and $\sigma^2$ is the noise power. $\text{PL}_n[t]$ is the average path loss in decibel (dB) between the UAV and node $n$ at a given time $t$ and is expressed as in \cite{sem_uav_cop}.

To focus on the downlink transmission between the BS and the UAV, we assume that the transmission of the UAV's state information at the end of the time slot $t$, from the UAV to the BS is ideal\footnote{This can be justified by the fact that the payload of such transmission is small and the bandwidth allocated to this communication is sufficiently high.}, i.e., free from packet loss or delay, as in \cite{goal_orient_sem_com_uav}. In contrast, we consider a downlink delay\footnote{This delay capture the time spent by the BS to calculate the commands.} $\Delta[t]$ for the transmission of control commands (i.e., the acceleration vector $\mathbf{a}[t]$) from the BS to the UAV. Upon receiving the command, the UAV is assumed to immediately execute it. 


\section{Problem formulation}

We denote by $b_n[t] \in \{0,1\}$ where $n \in \mathcal{N}$ and $t \in \{1,..,K\}$, the binary variable that equals $1$ if IoT device $n$ falls within the range of the UAV during time interval $t$, and $0$ otherwise.
Our objective is to maximize the average quality of the reconstructed images by collecting as many symbols as possible from the IoT devices over the time horizon $T$. To achieve this objective, we
optimize the UAV's acceleration matrix $\mathbf{A}=\{\mathbf{a}[t]\}_{t=1}^{K}$ generated by the BS over time. The problem can be formulated as follows:
\vspace{-0.09cm}
\begin{subequations}
\begin{align}
P1:&\max_{\mathbf{A}} \sum_{t=1}^{K} \left( \frac{\sum_{n=1}^{N} b_n[t] \phi_n[t]}{\sum_{n=1}^{N} b_n[t]} \right), \label{eq:objective}\\
\text{s.t.} & \sum_{t=1}^K b_n[t] \leq 1, \label{eq:once_visited} \quad \forall n \in \mathcal{N},\\
 & \mathbf{v}[t]=\mathbf{a}[t]\tau+\mathbf{v}[t-1], \; \forall t \in \{2,..,K\},\label{eq:velo}
 \\
&\mathbf{q}^\text{UAV}[t]=\mathbf{v}[t]\tau+\mathbf{q}^\text{UAV}[t-1],\; \forall t \in \{2,..,K\},\label{eq:posi}
\\
& \mathbf{q}^\text{UAV}[K] = \mathbf{q}^\text{fin}. \label{eq:traj_cons}\\
& |{a}^x[t]|\leq a^x_\text{max}, \; |{a}^y[t]|\leq a^y_\text{max}, \forall t \in \{2,..,K\}\label{eq:amax}.
\end{align}
\end{subequations}
Constraint \eqref{eq:once_visited} ensures that each IoT device is visited at most once during time horizon $T$\footnote{To ensure fairness, the UAV should not spend excessive time hovering over the same IoT devices.}. Constraints \eqref{eq:velo} and \eqref{eq:posi} describe the discrete motion dynamics of the UAV. Constraint \eqref{eq:traj_cons} guarantees that the UAV arrives at the final position $\mathbf{q}^\text{fin} = (x^\text{fin}, y^\text{fin}, h^\text{UAV})$ at the end of the $K$-th slot. Finally, constraint \eqref{eq:amax} ensures that the accelerations along the $x-$ and $y$-axes are bounded for all time intervals.


\section{Double Deep Q-Learning based Command and Control Approach}
To solve this challenging problem, we adopt a reinforcement learning (RL) approach \cite{mnih2013playingatarideepreinforcement, vanhasselt2015deepreinforcementlearningdouble}. The problem is sequential and involves making real-time decisions under the uncertainty of the channel, variable connectivity with IoT devices, and dynamic movement of the UAV. Traditional optimization techniques struggle to scale in such complex, high-dimensional environments with implicit dynamics governed by UAV motion. Reinforcement learning, and specifically Double Deep Q-Learning (DDQN), offers a practical and scalable solution.

\begin{figure*}
    \centering
    \begin{tabular}{ccc}
    \includegraphics[width=0.35\linewidth]{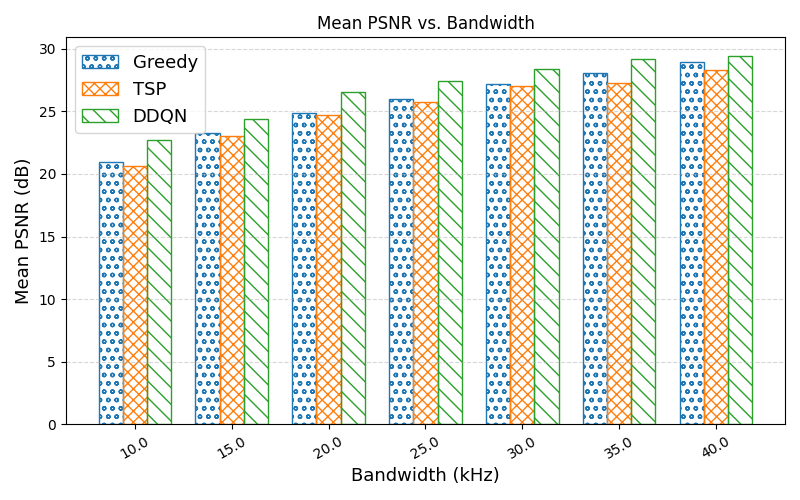} 
    
    &\hspace{-0.5cm}\includegraphics[width=0.35\linewidth]{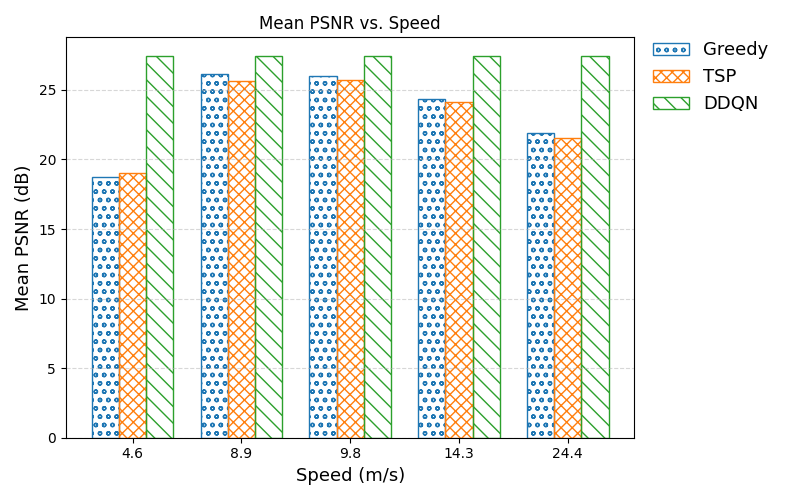}
    & \hspace{-0.5cm}\includegraphics[width=0.3\linewidth]{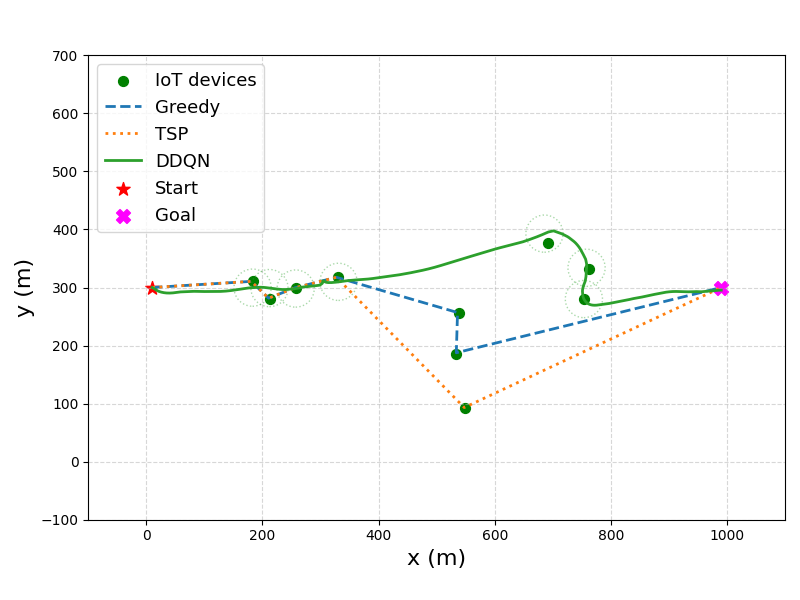}\\
    (a) Average PSNR vs. Bandwidth. &\hspace{-0.75cm}(b) Average PSNR vs. Velocity. &(c) Trajectory of UAV.
    \end{tabular}
    \caption{Performance of the proposed DDQN-based approach against benchmarks.}
    \label{fig:Comparison}
\end{figure*}
We consider a reinforcement learning agent deployed at the BS to control the UAV by generating its acceleration commands. The problem is modeled as a Markov Decision Process (MDP) defined by the tuple $(\mathcal{S}, \mathcal{A}, \mathcal{P}, \mathcal{R})$, where $\mathcal{S}$ is the set of states, $\mathcal{A}$ the set of actions, $\mathcal{P}$ the state transition probability function, and $\mathcal{R}$ the reward function. 
At the end of time slot $t$, the environment is in state $\mathbf{S}[t]$=\Big($\mathbf{q}^{\text{UAV}}[t],\mathbf{v}[t],(b_n[t])_{1\leq n\leq N},(\phi_n[t])_{1\leq n\leq N}, \Delta [t]\Big)$ which represents the UAV's position and velocity, the visited IoT devices, the corresponding quality of the collected images, and the remaining mission time. This state information is communicated to the BS by the UAV.
At the beginning of time slot $t+1$, the agent observes the current state $\mathbf{S}[t]$ and selects an action $\mathbf{a}[t+1]$. After the action is executed by the UAV, the environment transitions to a new state $\mathbf{S}[t+1]$ according to the transition probability $\mathcal{P}(\mathbf{S}[t+1]|\mathbf{S}[t], \mathbf{a}[t+1])$, and the agent receives a reward $r[t] \sim \mathcal{R}(\mathbf{S}[t], S[t+1])$. Accordingly, the reward at time $t \in \{1,..,K-1\} $ is: 
\vspace{-0.09cm}
\begin{equation}
    r[t] = r_c[t] + r_d[t] + r_g[t]
\end{equation}
The first reward term $r_c[t] = \frac{\sum_{n=1}^{N} b_n[t] \phi_n[t]}{\sum_{n=1}^{N} b_n[t] + \epsilon}$ ($\epsilon > 0$ is a tiny constant to avoid division by zero) is to encourage the UAV to collect data and to visit as many IoT devices as possible.
The term $r_d[t]$ is the reward term related to mission completion constraint. It encourages the UAV to arrive at its final destination within the allowed mission time $T$ \cite{final_pos}.
The term $r_g[t] > 0$ is a constant reward given when the UAV arrives at the final destination \cite{final_pos}.

To select the action $\mathbf{a}[t+1]$, the agent follows a policy $\pi$. Given the current state $\mathbf{S}[t]$, $\pi(\mathbf{a}[t+1]|\mathbf{S}[t])$ specifies the probability of choosing the action $\mathbf{a}[t+1]$. Given an initial state $s_1$, the goal is then to learn an optimal policy $\pi^*$ that maximizes the expected cumulative discounted reward.
To achieve this goal, we adopt DDQN, which reduces overestimation bias\cite{vanhasselt2015deepreinforcementlearningdouble}. This is done using two neural networks: an online network $\theta$ and a target network $\theta^{-}$. The online network $\theta$ is used to approximate the Q-values $Q(s,a; \theta)$ and select the best action for the next state. The target neural network $\theta^{-}$ is used for evaluating that action's value.
At each training step $t$, the algorithm maintains a replay memory $U$ that stores transition tuples $(\mathbf{S}[t], \mathbf{a}[t+1], r[t], \mathbf{S}[t+1])$. The replay memory stores past experiences, and once it has accumulated enough samples, the agent randomly selects a batch for the training of the online network. The loss function used to update the online network $\theta$ through gradient descent can be expressed as:
\begin{align}
    L(\theta) = &\mathbb{E}_{(\mathbf{S}[t], \mathbf{a}[t+1], r[t], \mathbf{S}[t+1]) \sim U}  \Bigl[ \Bigl (r[t] + \gamma Q \Bigl(\mathbf{S}[t+1], \\ \arg\max_{a} \nonumber & Q(\mathbf{S}[t+1],a;\theta);\theta^{-} \Bigr)  -Q(\mathbf{S}[t], \mathbf{a}[t+1]; \theta)\Bigr)^2 \Bigr].
\end{align}
The target network $\theta^{-}$ is periodically copied from the online network $\theta$ and kept fixed for a number of episodes.

{\color{red} In practice, constraint \eqref{eq:once_visited} is not enforced in the DDQN implementation, giving the agent more flexibility to decide how many time slots to remain within range of each device.}

\section{Simulation results}

To evaluate the performance of the proposed framework, we consider a $1000\times 600~\text{m}^2$ area in which $10$ IoT devices are randomly distributed. Each IoT device $n$ transmits DeepJSCC-encoded symbols using a transmit power of $P_n=1~\text{mW}$ over a bandwidth of $B_n =20~\text{kHz}$. The noise power is assumed to be $\sigma^2=10^{-9}~\text{W}$. The maximum number of time slots is $K=250$ and the duration of a time slot is $\tau = 0.5~\text{s}$. The UAV altitude is $h^\text{UAV}=100 ~\text{m}$ and the communication range is $D^\text{com} = 105~\text{m}$ and $r_g = 500$. 
The DDQN agent is a fully connected network of four layers. The online and the target networks have the same architecture consisting of three hidden layers with 128, 128, and 64 neurons, respectively. Each hidden layer is followed by a ReLU activation. The replay buffer size is 50000 and we used the Adam optimizer with a learning rate of $5\times 10^{-5}$.

The proposed DDQN approach is compared against two baselines: greedy and Traveling Salesman Problem (TSP). The greedy policy always directs the UAV toward the nearest unvisited IoT device within the mission horizon. In contrast, the TSP policy follows a predetermined global tour: at each time step, the UAV is directed toward the next device specified by the TSP solution. {\color{red} We used the OR-Tools routing library to solve the open-loop TSP (the starting point and the destination are not the same) \cite{ortools_routing}}. Both baselines operate with a fixed UAV velocity.

In Fig.~\ref{fig:Comparison}(a), we plot the mean PSNR as a function of the bandwidth allocated to the IoT devices  with a fixed velocity of $8.9~\text{m/s}$ for both greedy and TSP approaches. The proposed DDQN approach outperforms the benchmark schemes across all bandwidth values. At low bandwidth, greedy and TSP approaches suffer from degradation in reconstruction quality since only a limited number of symbols can be transmitted during each device visit. On the other side, DDQN learns to adapt the UAV's hovering time within communication ranges, which improves the image quality. As the bandwidth increases, the performance gap narrows, but DDQN still achieves the highest PSNR.

Figure~\ref{fig:Comparison}(b) illustrates the mean PSNR versus the UAV's velocities used in greedy and TSP approaches for a fixed bandwidth $B=25~\text{kHz}$. Unlike greedy and TSP, which operate with fixed velocities, DDQN optimizes the UAV's speed through adaptive C\&C acceleration commands. The results show that DDQN outperforms the benchmarks, particularly at very low and very high speeds. At low velocities, greedy and TSP cannot reach enough IoT devices within the mission horizon, which limits the number of collected symbols. At high velocities, both baselines force the UAV to leave the communication range too quickly, which results in incomplete symbol transmission and degraded reconstruction quality. DDQN mitigates these issues by balancing trajectory decisions to remain long enough within the range of communication of UAVs. Finally, Fig.\ref{fig:Comparison} (c) plots the trajectory of the UAV for the three approaches with a fixed velocity for greedy and TSP of $8.9~\text{m/s}$ and a bandwidth of $B=25~\text{kHz}$. The greedy policy relies on short-term paths  and the TSP policy follows a predetermined global path, ignoring communication delays and semantic data quality. In contrast, the DDQN trajectory adapts dynamically, allowing the UAV to linger around IoT devices and optimize the quality of collected data.

\section{Conclusion}

In this paper, we have presented a novel semantic‑aware UAV data collection framework that optimizes C\&C transmissions from the BS to the UAV. Our framework uses DeepJSCC at the IoT devices to encode the images into channel symbols and a Double Deep Q‑Learning agent at the BS to continuously adapt the UAV's trajectory for efficient data collection. By formulating the problem as a Markov Decision Process and explicitly accounting for downlink delays, our approach learns an adaptive flight policy that maximizes the average image quality.
Simulation results showed that the DDQN‑based approach outperforms both the greedy and TSP.

\clearpage
  \bibliographystyle{IEEEtran}
  \bibliography{bibliography}

@article{xu2023deepjointsourcechannelcoding,
  title={Deep joint source-channel coding for semantic communications},
  author={Xu, Jialong and Tung, Tze-Yang and Ai, Bo and Chen, Wei and Sun, Yuxuan and G{\"u}nd{\"u}z, Deniz},
  journal={IEEE communications Magazine},
  volume={61},
  number={11},
  pages={42--48},
  year={2023},
  publisher={IEEE}
}

@article{Bourtsoulatze_2019,
   title={Deep Joint Source-Channel Coding for Wireless Image Transmission},
   volume={5},
   ISSN={2372-2045},
   url={http://dx.doi.org/10.1109/TCCN.2019.2919300},
   DOI={10.1109/tccn.2019.2919300},
   number={3},
   journal={IEEE Transactions on Cognitive Communications and Networking},
   publisher={Institute of Electrical and Electronics Engineers (IEEE)},
   author={Bourtsoulatze, Eirina and Burth Kurka, David and Gunduz, Deniz},
   year={2019},
   month=sep, pages={567–579} }

@article{kurka2021bandwidthagileimagetransmissiondeep,
  title={Bandwidth-agile image transmission with deep joint source-channel coding},
  author={Kurka, David Burth and G{\"u}nd{\"u}z, Deniz},
  journal={IEEE Transactions on Wireless Communications},
  volume={20},
  number={12},
  pages={8081--8095},
  year={2021}
}

@ARTICLE{final_pos,
  author={Wang, Xueyuan and Gursoy, M. Cenk and Erpek, Tugba and Sagduyu, Yalin E.},
  journal={IEEE Internet of Things Journal}, 
  title={Learning-Based UAV Path Planning for Data Collection With Integrated Collision Avoidance}, 
  year={2022},
  volume={9},
  number={17},
  pages={16663-16676},
  doi={10.1109/JIOT.2022.3153585}}

@INPROCEEDINGS{sem_uav_cop,
  author={Wu, Yabin and Zhang, Fan and Xu, Chao and Wang, Xijun},
  booktitle={2023 IEEE 97th Vehicular Technology Conference (VTC2023-Spring)}, 
  title={Semantics-Aware Multi-UAV Cooperation for Age-Optimal Data Collection: An Adaptive Communication based MARL Approach}, 
  year={2023},
  volume={},
  number={},
  pages={1-5},
  doi={10.1109/VTC2023-Spring57618.2023.10200769}}

@inproceedings{xu2024semanticawarepowerallocationgenerative,
  title={Semantic-Aware Power Allocation for Generative Semantic Communications with Foundation Models},
  author={Xu, Chunmei and Mashhadi, Mahdi Boloursaz and Ma, Yi and Tafazolli, Rahim},
  booktitle={Proc. of IEEE Global Communications Conference (GLOBECOM)},
      year={2024}
}

@misc{liu2024diffusionbasedgenerativemulticastingintentaware,
      title={Diffusion-based Generative Multicasting with Intent-aware Semantic Decomposition}, 
      author={Xinkai Liu and Mahdi Boloursaz Mashhadi and Li Qiao and Yi Ma and Rahim Tafazolli and Mehdi Bennis},
      year={2024},
      eprint={2411.02334},
      archivePrefix={arXiv},
      primaryClass={cs.IT},
      url={https://arxiv.org/abs/2411.02334}, 
}

@ARTICLE{goal_orient_sem_com_uav,
  author={Wu, Wenchao and Yang, Yuanqing and Deng, Yansha and Hamid Aghvami, A.},
  journal={IEEE Transactions on Wireless Communications}, 
  title={Goal-Oriented Semantic Communications for Robotic Waypoint Transmission: The Value and Age of Information Approach}, 
  year={2024},
  volume={23},
  number={12},
  pages={18903-18915},
  doi={10.1109/TWC.2024.3424493}}

@article{xu2023taskorientedsemanticsawarecommunicationwireless,
  title={Task-oriented semantics-aware communication for wireless {UAV} control and command transmission},
  author={Xu, Yujie and Zhou, Hui and Deng, Yansha},
  journal={IEEE Communications Letters},
  volume={27},
  number={8},
  pages={2232--2236},
  year={2023}
}

@article{survey_uav_data_collection,
author = {Messaoudi, Kaddour and Oubbati, Omar Sami and Rachedi, Abderrezak and Lakas, Abderrahmane and Bendouma, Tahar and Chaib, Noureddine},
title = {A survey of UAV-based data collection: Challenges, solutions and future perspectives},
year = {2023},
issue_date = {Jul 2023},
publisher = {Academic Press Ltd.},
address = {GBR},
volume = {216},
number = {C},
issn = {1084-8045},
url = {https://doi.org/10.1016/j.jnca.2023.103670},
doi = {10.1016/j.jnca.2023.103670},
journal = {J. Netw. Comput. Appl.},
month = jul,
numpages = {44}
}

@article{chaccour2022dataknowledgebuildinggeneration,
  title={Less data, more knowledge: Building next-generation semantic communication networks},
  author={Chaccour, Christina and Saad, Walid and Debbah, Merouane and Han, Zhu and Poor, H Vincent},
  journal={IEEE Communications Surveys \& Tutorials},
  volume={27},
  number={1},
  pages={37--76},
  year={2024},
  publisher={IEEE}
}

@ARTICLE{licea2025reshaping,
  author={Licea, Daniel Bonilla and Silano, Giuseppe and El Hammouti, Hajar and Ghogho, Mounir and Saska, Martin},
  journal={IEEE Communications Magazine}, 
  title={Reshaping UAV-Enabled Communications with Omnidirectional Multi-Rotor Aerial Vehicles}, 
  year={2025},
  volume={63},
  number={5},
  pages={94-100},
  doi={10.1109/MCOM.001.2400421}}

@software{ortools_routing,
  title = {OR-Tools Routing Library},
  version = { v9.12 },
  author = {Vincent Furnon and Laurent Perron},
  organization = {Google},
  url = {https://developers.google.com/optimization/routing/},
  date = { 2025-02-17 }
}

@inproceedings{marnissi2024semantic,
  title={Semantic-aware resource allocation in constrained networks with limited user participation},
  author={Marnissi, Ouiame and Hammouti, Hajar EL and Bergou, El Houcine},
  booktitle={2024 IEEE Wireless Communications and Networking Conference (WCNC)},
  pages={1--6},
  year={2024},
  organization={IEEE}
}

@inproceedings{ndiaye2022age,
  title={Age-of-updates optimization for {UAV}-assisted networks},
  author={Ndiaye, Mouhamed Naby and Bergou, El Houcine and Ghogho, Mounir and El Hammouti, Hajar},
  booktitle={IEEE Global Communications Conference (GLOBECOM)},
  pages={450--455},
  year={2022},
  organization={IEEE}
}

@misc{qiao2024latencyawaregenerativesemanticcommunications,
      title={Latency-Aware Generative Semantic Communications with Pre-Trained Diffusion Models}, 
      author={Li Qiao and Mahdi Boloursaz Mashhadi and Zhen Gao and Chuan Heng Foh and Pei Xiao and Mehdi Bennis},
      year={2024},
      eprint={2403.17256},
      archivePrefix={arXiv},
      primaryClass={cs.IT},
      url={https://arxiv.org/abs/2403.17256}, 
}

@inproceedings{ndiaye2023muti,
  title={Muti-agent proximal policy optimization for data freshness in {UAV}-assisted networks},
  author={Ndiaye, Mouhamed Naby and Bergou, El Houcine and El Hammouti, Hajar},
  booktitle={IEEE International Conference on Communications Workshops (ICC Workshops)},
  pages={1920--1925},
  year={2023},
  organization={IEEE}
}

@inproceedings{vanhasselt2015deepreinforcementlearningdouble,
  title={Deep reinforcement learning with double {Q}-learning},
  author={Van Hasselt, Hado and Guez, Arthur and Silver, David},
  booktitle={Proceedings of the AAAI conference on artificial intelligence},
  volume={30},
  number={1},
  year={2016}
}

@misc{mnih2013playingatarideepreinforcement,
      title={Playing Atari with Deep Reinforcement Learning}, 
      author={Volodymyr Mnih and Koray Kavukcuoglu and David Silver and Alex Graves and Ioannis Antonoglou and Daan Wierstra and Martin Riedmiller},
      year={2013},
      eprint={1312.5602},
      archivePrefix={arXiv},
      primaryClass={cs.LG},
      url={https://arxiv.org/abs/1312.5602}, 
}

@ARTICLE{frombitstosemantics,
  author={Qin, Zhijin and Liang, Le and Wang, Zijing and Jin, Shi and Tao, Xiaoming and Tong, Wen and Li, Geoffrey Ye},
  journal={Proceedings of the IEEE}, 
  title={{AI} Empowered Wireless Communications: From Bits to Semantics}, 
  year={2024},
  volume={112},
  number={7},
  pages={621-652},
  doi={10.1109/JPROC.2024.3437730}}

\end{document}